# A Survey on Agent Workflow – Status and Future

Chaojia Yu
School of Computer Science
Sichuan University
Chengdu, China
yuchaojia82@gmail.com

Zihan Cheng
School of Computer Science
Sichuan University
Chengdu, China
czh051228@gmail.com

Hanwen Cui
School of Computer Science
Sichuan University
Chengdu, China
gmtd3328696811@gmail.com

Yishuo Gao
School of Computer Science
Sichuan University
Chengdu, China
gaoyishuo1@gmail.com

Zexu Luo
School of Computer Science
Sichuan University
Chengdu, China
luozexu535@gmail.com

Yijin Wang
School of Computer Science
Sichuan University
Chengdu, China
2024141520284@stu.scu.edu.cn

Hangbin Zheng
School of Computer Science
Sichuan University
Chengdu, China
zhenghangbin2006@gmail.com

Yong Zhao*
School of Computer Science
Sichuan University
Chengdu, China
yong.zhao@scupi.cn
*Corresponding author

*Abstract*—In the age of large language models (LLMs), autonomous agents have emerged as a powerful paradigm for achieving general intelligence. These agents dynamically leverage tools, memory, and reasoning capabilities to accomplish user-defined goals. As agent systems grow in complexity, agent workflows—structured orchestration frameworks have become central to enabling scalable, controllable, and secure AI behaviors. This survey provides a comprehensive review of agent workflow systems, spanning academic frameworks and industrial implementations. We classify existing systems along two key dimensions: functional capabilities (e.g., planning, multi-agent collaboration, external API integration) and architectural features (e.g., agent roles, orchestration flows, specification languages). By comparing over 20 representative systems, we highlight common patterns, potential technical challenges, and emerging trends. We further address concerns related to workflow optimization strategies and security. Finally, we outline open problems such as standardization, and multi-modal integration—offering insights for future research at the intersection of agent design, workflow infrastructure, and safe automation.

## I. Introduction

In the age of artificial intelligence, automation is no longer a mere engineering convenience but a shared aspiration. Building autonomous systems becomes an efficient path toward discovering the paradigm of intelligence.

Among various efforts, the emergence of large language models (LLMs) has revolutionized natural language understanding and decision-making, demonstrating remarkable capabilities in reasoning, planning, and tool-use coordination.

Researchers have begun exploring how to grant LLMs more autonomy in decision-making and task execution. For example, Auto-GPT is a product of an experimental project developed to make the use of GPT-4 autonomous [1].

To solve a problem or make a decision, we naturally follow some order by planning in advance, then taking a sequence of actions to complete the task. In human cognition, it is natural to represent problem-solving as a step-by-step procedure—a workflow that clarifies "what happens next" in a structured manner. Equipped with the capability of agents, we have a big step from manually pre-defined workflow. However, as more agent workflows are introduced by major companies [2][3], the absence of a unified workflow framework is becoming increasingly clear. Individual agents—no matter how powerful—operate like isolated units, unable to cooperate effectively or adapt to dynamic requirements. In this context, workflow is not only a task execution tool but serves as the backbone of the emerging AI ecosystem, orchestrating agents across roles, capabilities, and modalities. Ultimately, the goal of agent workflow research is to enable agents to operate fully autonomous in real-world scenarios involving complex, multi-step task.

This survey provides a systematic introduction to agent workflows and offers a comparative analysis of their capabilities, architectures, and underlying mechanisms. It aims to help readers understand the current status and future directions of agent workflows.

The remainder of this survey is structured as follows. Section 2 reviews the background. Section 3 presents an overview of common frameworks of agent workflows, focusing on architecture, specification and workflow management mechanisms. Section 4 explores a comprehend comparison for current agent workflows. Section 5 lists several workflow-level optimization strategies. Section 6 highlights major application domains. Section 7, 8, 9 discuss security issues, limitations and future work, provide conclusion, respectively.

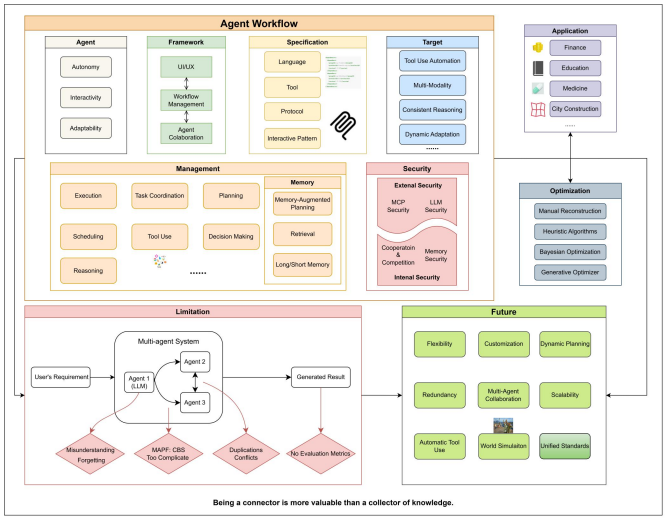

Fig. 1. Overview of the Survey

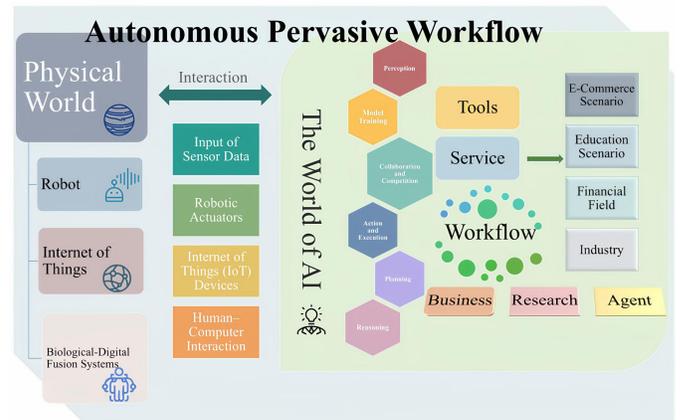

Fig. 2. The World of Agent Workflow

## II. BACKGROUND

### A. Definition

Agent: Agents are systems where LLMs dynamically direct their own processes and tool usage, maintaining control over how they accomplish tasks [4].

The core features of agent is autonomy, interactivity and adaptability. A system can reasonably be regarded as an agent rather than a static LLM wrapper once we validate these features.

Workflow: A workflow is a system for managing repetitive processes and tasks which occur in a particular order [4].

A workflow can include operations such as asking the user for information, invoking tools, and responding to the user, which can be represented as a series of nodes or stages. One common way to represent and organize is to use a directed graph, the nodes represent distinct decision point, and the directed edges represent temporal or dependency relationships between nodes. To combine, the language model operates within an existing workflow by following a predefined process to accomplish tasks.

### B. Component of Agent

An agent has two main parts: The brain (AI model), usually LLM, which is for reasoning, planning and reflection. The body of an agent refers to its built-in capabilities (e.g., memory) and external tools used to complete actions.

### C. Multi-agent System (MAS)

Multi-agent systems (MAS) evolve from single-agent execution to collaborative interaction and ultimately to orchestrated work-flows. A single agent has simple attributes that can solve a particular inference task. However, complex problems require extensive collaboration and collective intelligence, which lead to Multi-agent System. The core advantage of MAS lies in its distributed decision-making and problem-solving capabilities [6].

### D. Evolution of Agent Workflow

The evolution of workflow can be divided into these four stages.

i. Business Process Management
ii. Data-driven Science and Research Workflow
iii. Agent Workflow
iv. Autonomous Pervasive

The traditional non-AI framework is usually pre-defined with hard-coded rules, hints and tool chains. The workflow is supported by designing the script manually, following a particular tool orchestration. This includes business-centric automation, research prototypes for cognition.

With the development of technology, agents are increasingly embedded into workflows to assist or automate decision-making. The agent workflow is defined as several agents follow the order to make a sequence of decisions, using available tools through interaction with the environment. To make sure LLM Agents follow an effective and reliable procedure to solve the given task, workflows are usually used to guide the working mechanism of agents [5]. To dealing with more complicated tasks, Thus we need to define complicated workflow.

Ultimately, the target is to build agents that are truly autonomous—not following any pre-set instructions but only general prompts, just like "aware" of its own process. We call it auto-pervasive agents that continuously act, reason, and adapt within real-world environments.

## III. FRAMEWORK OF AGENT WORKFLOWS

To achieve complicated workflow we must have a unified framework or paradigm.

### A. Multi-layer Architecture

The typical agent workflow architecture includes 3 layers.

1) UI/UX: Provides the interface layer for users to interact with agents intuitively.

2) Workflow Management: Coordinates the execution of tasks through structured processes, interacts with two other layers.

3) Agent Collaboration: Enables multiple agents [7] to cooperate, communicate, and delegate subtasks to achieve complex goals collectively.

*B. Roles of Agent*

Multi-agent workflows assume diverse roles for agents based on the system's needs and coordination strategies. Typical roles include:

Planner: Responsible for decomposing tasks and assigning responsibilities (e.g., Commander in AutoGen2 [8]).

Executor: Carries out specific subtasks or tool calls (e.g., Coder).

Parser/interpreter: interprets external data (e.g., input queries, file formats, code).

Critic/Reviewer: evaluates results or provides feedback. Memory Manager, Communicator, etc.

For example, in the context of simulating the operations of a school, appropriate roles would include teachers, students, and the principal.

*C. Specification*

Currently, AI agents developed by different vendors and research institutions often adopt disparate architectures, interface standards, communication protocols, and data formats, making it difficult for agents to interoperate directly. We urge to set a standardized specification and expect to solve the following questions like:

a. In which form can we represent workflows precisely?

b. How to control the behavior of LLMs effectively?

1) Language

- Prompt: Natural language instead of formal languages. This makes workflows highly flexible and human-readable, but also less standardized and harder to validate.
- Modeling Workflow: Traditional formal workflow languages, such as BPML (Business Process Modeling Language) and XPDL (XML Process Definition Language). Business Process Execution Language for Web Services (BPEL4WS) [9][10].define task nodes, state transitions, and interface specifications to formally describe the structure and execution logic of a workflow. Swift [74], VDL [75] use a functional-flavored scripting language to concisely describe large scale scientific workflows.
- Programming: python, etc.
- Declarative configuration: YAML, JSON, etc.

2) Tool: Tool-backed agents have the capability to execute external tools via code execution or function execution [3]. For Example, the systems have retrieval tools like vector search engines, web searcher, computation and reasoning tools like calculators, code interpreters, knowledge and query tools like databases, Q&A systems, APIs and function calling Interfaces like OpenAI function calling, toolbox Interface (e.g., AutoGPT ToolManager) [11]. This integration enables LLMs to access real-time knowledge and perform specialized operations [12].

3) Protocol:

**General-purpose service Protocols**: REST/HTTP that used in most agent tool calls and API wrapping.

WebSocket, etc. for real-time, bidirectional communication. These protocols are based on function-calling.

**Agent-native Protocols**:

Model Context Protocol (MCP): an open protocol that enables seamless integration between LLM applications and external data sources and tools [28].

Agent Network Protocol (ANP): an open source protocol for agent communication which aims to become the HTTP of the agentic web era. The vision is to define how agents connect with each other, building an open, secure, and efficient collaboration network for billions of intelligent agents.

4) Interactive Pattern: This specification mainly discuss how agents perceive environmental information to acquire knowledge and experience.

a. Interacting with **environment**. The components are environment, sensor, executor and effector, following the process of observation, action and feedback.

b. Interacting with **other human users**. Mainly through dialogues, following the procedure of "human-in-loop".

c. Interacting with **self or other agents**. These are discussed in the scope of reflection and MAS.

The targets are presented in Fig. 1.

*D. Workflow Management*

Since workflow management is a broad topic, we present key taxonomies and important aspects here.

Workflow Management System is a system that defines, creates and manages the execution of workflows through the use of software, running on one or more workflow engines, which is able to interpret the process definition, interact with workflow participants and, where required, invoke the use of tools and applications [10].

1) Workflow Mode:

a. Chain Workflow: Decomposing a general task into a sequence of steps, where each step depends on the output of the previous one.

b. Parallelization Workflow: Executing multiple tasks or processing multiple datasets simultaneously, where the tasks are independent of each other.

c. Routing Workflow: Dispatches tasks to specialized processors based on the type or characteristics of the input, particularly when task processing depends on input-specific features.

d. Orchestrator–Workers: A central LLM acts as the orchestrator, decomposing complex tasks and delegating subtasks to specialized worker agents.

e. Evaluator–Optimizer: One LLM generates responses while another evaluates their quality and provides feedback for refinement. This setup supports self-improvement and iterative optimization [13].

2) Workflow Execution: Workflow execution defines how an agent workflow is triggered, scheduled, and terminated. It determines whether a task proceeds in a fixed procedural order (static workflow), or adaptively changes based on agent decisions (dynamic workflow).

Driven by an execution engine that follows a looped pattern is a popular structure [1][14][74]. We also introduce execution units to help execution. For instance, Agent-as-a-Node means each node in the workflow is an autonomous agent responsible for a task. Agent-as-a-Planner means a single or few central agents dynamically plan the workflow path and assign execution to sub-modules.

This section is also highly related to task coordination, scheduling, tool use.

3) Problem Solving: Problem solving represents the holistic execution behavior of agent workflows, encompassing planning, acting, interacting, and adapting in pursuit of a goal. Problem solving permeates every stage of the workflow.

Representative works are not limited to planning alone: ReAct (planning, tool use), Tree-of-Thought (planning, memory, exploration) [15]. Typical application scenarios include software development and beyond.

4) Planning: Agent Workflow Structure Basically we follow 5 steps: perception, reasoning, decision making, action execution, feedback and learning.

The following "nodes" form a logical chain. **Discovery** refers to the agent's ability to autonomously identify unknown goals. **Task decomposition** dynamically decomposes complex tasks into smaller subtasks and assigns each to a specifically generated sub-agent, thereby enhancing adaptability in diverse and unpredictable real-world tasks. For example, researchers propose a multi-agent framework based on dynamic task decomposition and agent generation [11] with a particular focus on travel planning. Task decomposition generates a sequence of actions to achieve specific goals by a sequence of subtasks which serves as an intermediate state guiding the agent in subsequent steps. **Reasoning** typically involves decomposing complex questions into sequential intermediate steps, also known as chains before producing the final answer. It is currently heavily rely on Chain-of-Thought (CoT) to deal with the tasks like multi-hop question answering and fact verification. One representative approach is Reasoning via Planning (RAP) [16]. LLMs can serve as virtual domain experts to assist in specialized reasoning tasks, such as extracting causal orderings from variable descriptions to support causal inference [17]. **Action** is the next step. Considering decision-making process of agent workflow, the action at a given time is decided by all past actions and observations up to the time, and G serves as the guide. **Reflection** mainly focus on handling upstream failures, emergence of new information, and environmental changes, also when planning and acting in more complex environments. SayCan proposes predicting actions which are subsequently filtered through an affordance model [18]. Inner Monologue introduces a self-feedback mechanism to enhance reasoning and decision-making [19]. A typical scenario is coding as the generated code may contain errors or fail to meet user requirements.

In workflow management, memory plays a significant role in retrieval and storage.

*E. Illustrative Example*

LangChain is a framework [20] for developing applications utilizing large language models, and its goal is to enable developers to conveniently utilize other data sources and interact with other applications. The architecture of LangChain is composed by the following key modules:

**Model**: It provides the interfaces of different LLMs for users to use and also offers a set of standard message classes.

**Prompt Template**: It provides multiple classes and functions to facilitate the creation of prompt templates, as well as some tools for using them.

**Memory**: Langchain allow to store the conversation context, supporting both short-term and long-term memory, which enhance the consistency of the conversation.

**Chains**: Which allows the combination of different models and stages for the complex workflow of the AI agents. For example, one of the most common chains, the LLMChain, which combine the Prompt Template, Model and the Guardrails to obtain the response of the model.

LangChain has a variety of built-in tools and also provides toolkits to make it convenient for users. Meanwhile, users can register their own python functions as tools. Regarding the tool invocation, it supports the format of OpenAI functions and the output in JSON. This makes the return values of agents more standardized.

When the Agent gets a user request, it starts a think-and-act loop. First, the LLM reads the request and the current context. Then it decides whether to answer directly or use a tool. If it chooses to use a tool, it outputs a specific format with the tool's name and parameters. LangChain then runs the tool and gives the result back to the LLM. The LLM takes this result, updates its thinking, and decides what to do next. This back-and-forth keeps going until the LLM is ready to give the final answer. In LangChain, this process is called the ReAct agent pattern, which means the model "thinks" while taking "actions". Later it had iterations and variants such as LangChain Expression Language, LangGraph [21].

## IV. COMPARATIVE ANALYSIS OF AGENT WORKFLOW SYSTEM

The first perspective is to compare the capabilities across 24 agent workflow systems. The comparison is shown in Table 1.

### A. Metrics Description

- Planning: Do the system have the ability to independently plan task processes.
- Tool Use: Whether the agent can call external tools such as APIs, calculators.
- Multi-agent: Whether the system supports multiple agents working collaboratively.
- Memory: Whether the agent includes an explicit memory mechanism.
- GUI: Whether agents can interact with graphical interfaces.
- API: Whether agents interact with external systems through structured API calls, such as Function Calling.
- Self-Reflect: Indicates whether the agent has the ability to self-evaluate or reflect, a typical feature is looped procedure.
- Custom Tools: Whether the framework allows users to integrate or define new tools.
- Cross-platform: Whether the system can be deployed across multiple platforms, the platform indicates distinct operating systems, web or mobile integration, etc.
- Open Source: Whether the project is open-source and the source code is publicly available.
- Year: The release or open-sourcing year of the system or the relevant essays.

| | |
|---|---|
| √ | Support |
| × | Not Support |
| ◐ | Partially Support |
| √* | Only Support Specific API |
| ○ | Unspecified |

### B. Interpretation and Scope Clarification

1) We also determine Whether the API calling only support specific ones. For example many products of OpenAI only allow to call the API of Open AI. If so, append "*". For the explanation, if multiple models are supported, it will be necessary to unify the interfaces, optimize the prompt words, and control the context, all of which will increase the cost.

2) For the metric "Memory", it emphasizes context management and state maintenance of multiple rounds of historical conversations.

3) Compared to API-based agent, a distinctive form of tool use arises when agents operate in graphical user interface (GUI) environments. Rather than invoking back-end APIs or structured function calls, GUI-based agents interact with applications through human-like operations such as mouse clicks, keyboard inputs, or visual element selection [22].

The second perspective is to compare the architecture and mechanism across 24 agent workflow systems. It is shown in Table 2.

### C. Metrics Description

- Agent Roles: Specifies whether the framework supports role-specific agents.
- Flow: Describes how the workflow execution is structured. Data Flow focuses on how data moves between modules, while Control Flow emphasizes the logic or sequence of execution (e.g., loops, branches). Some system are mixed but dominated by one of them.
- Representation: The formalism used to represent the workflow or task plan.
- Language: Indicates the language or interface used to help define workflow.
- Protocol: Refers to the communication mechanism between agents or modules, including newer protocols like MCP (Model Context Protocol) or custom APIs.
- Deployment: The primary deployment mode supported by the system.

### D. Interpretation and Scope Clarification

1) Data flow represents a workflow as a data state machine, where context is passed between nodes via "state objects." The execution order does not need to be predefined, as node selection is determined by data content and transition conditions.

2) Language is not targeting at modeling language but also include lightweight data description languages. The language are sometimes domain-specific. Unlike traditional workflow systems relying on formal specification languages like BPEL or BPMN, most LLM-driven agent frameworks adopt lightweight declarative or prompt-based descriptions. This reflects a broader trend

Table 1. Comparison of Capabilities Across Agent Workflow Systems

| System | Planning | Tool Use | Multi-agent | Memory | GUI | API | Self-Reflection | Custom Tools | Cross-Platform | Open-source | Year |
|---|---|---|---|---|---|---|---|---|---|---|---|
| AgentUniverse [54] | √ | √ | √ | √ | √ | √ | ◯ | √ | √ | √ | 2023 |
| Agentverse [55] | √ | √ | √ | √ | ◐ | √ | √ | √ | √ | √ | 2023 |
| Agno [56] | √ | √ | √ | √ | × | √ | √ | √ | √ | √ | 2024 |
| AutoGen [3] | √ | √ | √ | √ | × | √ | √ | √ | ◐ | √ | 2023 |
| CAMEL [57] | √ | √ | √ | √ | × | √ | √ | √ | √ | √ | 2023 |
| ChatDev [58] | √ | √ | √ | √ | × | × | ◐ | √ | √ | √ | 2023 |
| Coze [59] | √ | √ | √ | √ | √ | √ | ◯ | √ | √ | √ | 2024 |
| CrewAI [60] | √ | √ | √ | ◯ | × | √ | √ | √ | ◐ | √ | 2024 |
| DeepResearch [61] | √ | √ | ◯ | √ | √ | √* | ◐ | √ | √ | × | 2025 |
| Dify [62] | √ | √ | × | √ | × | √* | ◯ | √ | √ | √ | 2023 |
| DSPy [63] | √ | √ | ◯ | √ | √ | √* | √ | √ | √ | √ | 2023 |
| ERNIE-agent [64] | √ | √ | √ | ◯ | √ | √ | × | √ | √ | √ | 2024 |
| Flowise [65] | √ | √ | × | √ | × | √ | × | √ | √ | √ | 2023 |
| LangGraph [66] | √ | √ | √ | √ | × | √ | √ | √ | ◐ | √ | 2023 |
| Magnetic-One [67] | √ | √ | √ | √ | √ | √ | √ | √ | ◐ | √ | 2024 |
| Meta-GPT [2] | √ | √ | √ | ◐ | × | √* | × | √ | √ | √ | 2023 |
| n8n [68] | √ | √ | × | √ | √ | √ | × | √ | √ | √ | 2019 |
| OmAgent [69] | √ | √ | × | √ | × | √ | √ | √ | ◯ | √ | 2024 |
| OpenAI Swarm [70] | √ | √ | √ | × | × | √ | × | √ | √ | √ | 2024 |
| Phidata [71] | √ | √ | √ | √ | × | √* | × | √ | √ | √ | 2024 |
| Qwen-agent [72] | √ | √ | √ | √ | √ | √ | √ | √ | √ | √ | 2024 |
| ReAct [14] | √ | √ | × | × | × | √ | × | √ | × | √ | 2022 |
| ReWoo [23] | √ | √ | × | × | × | √ | × | √ | × | √ | 2024 |
| Semantic Kernel [73] | √ | √ | × | √ | × | √* | × | √ | √ | √ | 2023 |

toward flexibility and rapid iteration in intelligent agent systems.

*E. Performance Snapshots*

We provide a closer look at the capabilities of several representative agent workflow systems.

The system n8n, an non-LLM-driven agent, exhibits behaviors that are highly similar to those of an agent workflow system in terms of process automation, tool integration, and task scheduling, which is a typical "no-code agent execution shell".

ReWoo [23] proposes a composable, modular, and low-token-cost approach to reasoning workflows. Instead of fol-lowing the ReAct-style step-by-step tool-feedback prompting, ReWoo adopts a "reasoning-without-observation" strategy—planning the tool calls in advance and then executing them collectively. This resembles a "planner + executor" pattern, with a particular focus on reasoning efficiency and low coupling between components.

Agno closely aligns with the ReAct and AutoGen paradigm and is a lightweight framework for building multi-modal agents. Its "team mode" introduces explicit multi-agent workflow characteristics, enabling structured collaboration among agents.

ReAct [14] is a prompting framework that interleaves natural language reasoning traces with task-specific actions, enabling LLMs to update plans, handle exceptions, and interact with external environments in a structured loop. It achieves superior performance and interpretability across QA, fact-checking, and decision-making tasks, and forms the foundation for many modern agent workflows. It is then widely accepted by systems like LangChain Agent, AutoGPT (Action Planner), WebGPT, HuggingGPT, DSPy, etc.

AutoGen has been used to construct a structured multi-agent workflow for coding-based reasoning tasks, such as interpreting optimization results in OptiGuide. In this system, the user submits a question to a Commander agent, which coordinates two sub-agents: the Writer, responsible for code generation and explanation, and the Safeguard, responsible for checking code safety. The Commander acts as the central controller, executing code (e.g., via Python), relaying results to the Writer for interpretation, and handling exceptions by routing debugging information back when necessary. This back-and-forth process may iterate multiple times until the task is completed or times out [3].

Table 2. Comparison of Architectures and Mechanisms Across Agent Workflow Systems

| System | Agent Roles | Flow | Representation | Language | Protocol | Deployment |
|---|---|---|---|---|---|---|
| AgentUniverse | PEER, DOE | Control | Pattern Factory | python, YAML | ○ | Local(mainly) |
| Agentverse | Expert, Decider, etc | Control | Stage-based | python | Self-defined | Local or Specific Environment |
| Agno | Single/Team | Mixed | Trace | python | API-based | Local/Cloud |
| AutoGen | Commander, Worker, Critic | Control | DAG | python | Function Schema | Local |
| CAMEL | Planner, Executor, etc | Control | Modular Graph | YAML-based | MCP | Local/Web |
| ChatDev | CEO/CTO/CPO/Programmer | Control | DAG-like | python | ○ | ○ |
| Coze | Conversational | Control | Node-based | ○ | API-based | Web/Mobile/API Endpoint |
| CrewAI | Planner, CrewMember | Control | Plan Graph | python DSL | Function Schema | Local CLI |
| Deep Research | Searcher, Analyzer, etc | Mixed | Semantic Plan Trace | ○ | Internal | OpenAI Only |
| Dify | NA | Control | Prompt chain | JSON | Function Schema | Saas/Local |
| DSPy | Planner, Retriever, etc | Mixed | Modular Graph | python | API-based | Local/Distributed |
| ERNIE-agent | Implicit | Mixed | Flowchart | python | ○ | Local CLI |
| Flowise | NA | Data | DAG | JSON | Langchain | Web/Docker/Local |
| LangGraph | By Node | Mixed | DAG | python SDK | Langchain tool protocal | CLI/SDK |
| Magnetic-One | Orchestrator, Coder | Control | DAG | python SDK | AutoGen-Chat based | Local |
| Meta-GPT | PM, Engineer, etc | Control | Class | python | ○ | Local CLI |
| n8n | By Node | Control | Flowchart | javascript | Webhook, OAuth, REST API | Cloud/Docker/Local |
| OmAgent | Planner, Retriever, etc | Control | Text Plan | python | ○ | Specific System |
| OpenAI Swarm | Worker, Router | Mixed | Encapsulated | python, YAML | NA | Local |
| Phidata | Team | Control | DAG-like | python DSL | ○ | Web/Local |
| Qwen-agent | Self-defined | Control | Code | python | MCP | Local/Cloud |
| ReAct | NA | ○ | Step List | ○ | NA | NA |
| ReWoo | NA | Mixed | Script | python | Self-defined | Local |
| Semantic Kernel | Implicit | ○ | DAG | python, YAML | Function Schema | Local/Cloud/Saas |

## V. OPTIMIZATION

Although there has been extensive research on agent optimization techniques—including learning strategies, reward shaping, and system efficiency improvements [24], few works have explicitly focused on optimizing the agent workflow itself, such as scheduling, coordination structure, or workflow representation.

From the view of agent workflow, our direction are such as multi-task scheduling, multi-objective optimization, resource-aware allocation, and even reinforcement learning-based adaptive schedulers. Reinforcement-based strategies are particularly promising in dynamic environments, where agents must make decisions under uncertainty and learn optimal policies for long-term efficiency. For example, token usage is a significant optimization target. Each invocation of a large language model incurs token consumption, and the cumulative cost can quickly escalate. Therefore, there is a necessary trade-off between model performance and latency and cost control.

Optimization strategies on agent workflow can be roughly divided into 4 categories:

**Manual Reconstruction**: For workflow with a few agents, optimization can be applied manually after careful analysis and comparison between different strategies. However, workflow with more agents can hardly be handled due to complex relationships and implicit causality.

**Heuristic Algorithm**: Heuristic algorithms enable optimizers to handle workflows that are implicit and intricate, largely increasing the efficiency of workflows without having to check each detail. Besides, these algorithms are naturally discrete, enabling them to handle discrete models. However, these heuristic approaches are easy to fall into local optima, and their requirement of manually set parameters leads to huge difference in optimization efficiency even with the same algorithm [25].

**Bayesian Optimization**: Bayesian optimization can be used to optimize agent workflows due to its efficiency in searching in discrete space [26]. It greatly improved the efficiency of small size workflow and successfully achieved multi-objective optimization. However, its effect on larger workflows still needs to be verified.

**Generative Optimizer**: Some researchers turn to LLMs to optimize workflows, by describing to LLMs the whole workflow to generate suggestions, and adding suggestions to the workflow, giving it back to LLMs for more suggestions [27]. These algorithms have great adaptability, while supporting scalar feedback, natural language and error messages for optimization. Though due to its dependency on LLM, non-text parameters can hardly be optimized, and the performances on stateful functions and distributed workflows are not inline with expectations.

## VI. APPLICATION

Currently, agent workflows have been applied across various fields, including healthcare, urban, finance, education, and law.

**Healthcare**: Gao, S. et al. [28] developed TxAGENT, a precision medicine therapeutic agent based on a multi-modal adaptive model, which employs multi-step reasoning and real-time biomedical knowledge retrieval to dynamically generate personalized treatment plans for patients. Min, R. et al. conducted a study on a medical question answering system leveraging LangChain and large language models (LLMs), utilizing a dynamic tool selection mechanism to intelligently switch between knowledge graphs and search engines, thereby improving accuracy.

**Urban Planning**: Ni, H. et al. [29] proposed an multi-agent LLM-based cyclical urban planning framework, which employs automated profiling generation and memory reflection mechanisms to achieve a dynamic closed-loop process of planning, simulation, and optimization.

**Finance**: Yang, H. et al. [30] proposed FinRobot—an open-source AI agent platform for financial applications. The platform enables collaborative work among modular agents, and utilizes Chain-of-Thought (CoT) technology to simulate the reasoning process of human analysts, decomposing complex financial problems into logical steps, and optimizing task execution through dynamic scheduling. Han, X. et al. [31] optimized a multi-agent collaborative system to enhance investment analysis in financial research, leveraging agent collaboration architectures to improve accuracy.

**Education**: Morales-Chan, M. et al. [32] developed an AI agent workflow using the LangChain framework and OpenAI API to provide personalized feedback for massive open online courses (MOOCs). Jiang, Y. H. et al. [33] centered their research on multi-agent systems in education, proposing a von Neumann multi-agent system framework that simulates human cognitive processes, enabling task deconstruction, self-reflection, and the management of short-term and long-term memory to support complex information integration and personalized teaching.

**Law**: Yue, S. et al. [34] studied the Multi-agent Legal Simulation Driver (MASER), which generates scalable synthetic data by simulating interactive legal scenarios, providing a rich training resource for legal AI.

Scene Customization refers to the process of adapting a general-purpose agent workflow framework to meet the requirements of specific application domains—such as academic research assistants, financial analysts, programming companions, or DevOps engineering assistants. This involves customizing task flows, defining specialized agent roles, selecting and integrating domain-specific tools, and designing appropriate feedback mechanisms. For instance, a research assistant may emphasize literature retrieval, citation formatting, and document summarization; a programming agent focuses on code generation, debugging, and API usage; while a DevOps agent is tailored for deployment workflows, log analysis, and system health monitoring.

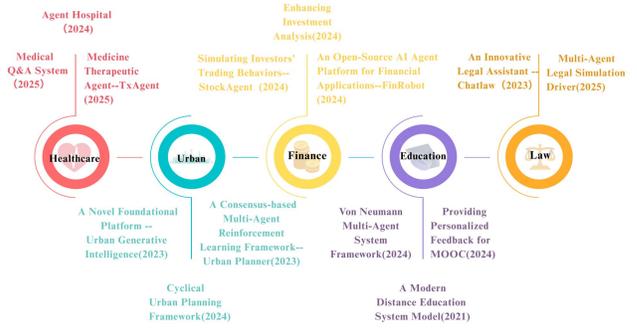

Fig. 3. Applications of Agent Workflow in Various Fields

## VII. SECURITY

The security issues associated with AI agent workflows have gained prominence due to their rapid development. These security issues can be broadly divided into two categories: internal security and external security. Internal security includes memory security and agent cooperation and competition, while external security primarily involves the interaction between the agents and external resources, such as the Model Context Protocol (MCP) [40] and security issues of large language models (LLM), as discussed in this article.

### A. External Security

Recently, Invariant Labs discovered a critical vulnerability in the MCP that enables "Tool Poisoning Attacks" [41]. In addition, there are certain problems with MCP servers as well as certain security threats.

1) Tools: The common attacks on agent tools include hidden Instructions [42], in which malicious prompts hidden in tool descriptions to steal user data. Rug pulls in which attackers alter tool descriptions after user authorization (e.g., in MCP). Tool name collision in which tools with conflicting names cause security risks during setup. Slash command overlap: Similar commands across tools may lead to execution conflicts and can be exploited.

2) MCP server: There are three common ways: Some malicious entities register with names similar or identical to legitimate MCP servers, deceiving users into installation [43]. Secondly, during the update phase, there are still some issues that may pose a threat to AI agents accessing the MCP. As it is a source project maintained by the community, it is necessary to manage official packages and standardize packaging formats to ensure timely updates [43]. In cross-server attacks [41], attackers can execute attacks by using shadowing tool descriptions. Malicious servers can modify descriptions of trusted tools.

3) LLM: Malicious inputs from users may lead to model contamination, affecting the integrity of the AI Agent [44]. Adversarial data poisoning attacks pose a significant threat to LLM-based AI Agents; seemingly harmless prompts, when concatenated, may contaminate the model. Chat records generated through natural user interaction may also inadvertently contaminate the model. The privacy issues include that there may be a risk of privacy leakage during the use of the AI Agent. When processing user prompt data for chatbots, there have been some confidentiality issues with LLMs [44][45]. The AI Agent often requests personal information when utilizing tools, which can be easily remembered through chat records, making the AI Agent susceptible to data extraction attacks.

*B. Inner Security*

In the process of the AI agent workflow, usually we do not just use one agent. Therefore, in a multi-agent system, we need to consider the issues of cooperation [46] and competition [47] among different agents. In addition, regarding the memory part, since it involves large language models and user privacy, it often becomes an entry point for attacks by malicious users.

1) MAS: In MAS collaboration the threats include covert collusion, illusion amplification, spread of misinformation, malicious attacks. For competition threats, malicious competitive behaviors, ethical concerns is found.

2) Memory: Memory security in AI agents can be divided into two parts: short–term and long-term. Short-term memory may encounter capacity limitations, leading to information loss. Asynchronization can also disrupt data flow. Long-term memory faces various threats. Poisoning attacks can corrupt stored data. Privacy breaches may expose sensitive information [48]. Additionally, there are generation threats that can lead to false outputs. Mitigating these risks is essential for reliable AI operation [42].

## VIII. LIMITATIONS AND FUTURE DIRECTIONS

We list several specific limitations here along with a limitation in a path finding strategy. Also we state the profound limitation for the research and industry field and give some future direction.

*A. Specified Limitations*

**Lack of Environmental Feedback**: The current system does not consider incorporating environmental feedback into the Automated Prompt Optimization (APO) process. Adding a feedback mechanism will enhance the system's interactivity and error-correction capabilities.

**LLMs Function Limit**: Some agents currently rely on large language models (LLMs) to collect user information and understand user needs. However, existing LLMs often forget or misinterpret previous continuous, progressive requests from users, leading to inaccurate information transfer from the source, which severely impacts the final results [49]. At the same time, LLM's feedback is directly related to the user's language expression ability. If the user is unable to describe their goal precisely, the LLM cannot infer the user's potential needs, causing the final results fail to meet the user's initial expectations, leading to user's dissatisfaction with the agent's output.

**Lack of Evaluation Metrics for Agents**: Current agents lack unified standards for assessing their quality. Open-source platforms and programs such as EvalAI [50] and DevAI are already available for evaluating AI systems, but they still overly focus on the agent's final output, neglecting a detailed analysis of the step-by-step working principles of the agent. Additionally, existing evaluation standards have limited testing scope, mostly focusing on specific scenarios defined by the testers, which doesn't generalize well to broader AI development contexts and fails to fully reflect the actual usage value of agents. Furthermore, there is a significant amount of human evaluation involved in the evaluating process, which is too subjective and unreliable as a unified judgment standard.

**Lack of Category Diversity**: There is a gap in the classification of agents within the agent development field. Classifications based on functionality or underlying tools are missing, which is detrimental to the future development of multi-agent systems built upon existing agents.

**Duplication, Redundancy, and Conflict**: In MAS, when multiple agents handle complex tasks concurrently (especially in multiple LLM systems), the lack of efficient multi-objective optimization and scheduling mechanisms leads to unnecessary and useless redundant information. As a result, the system needs to perform additional checks to remove duplicates, which consumes extra server resources and slows down progress. When agents exchange data, conflicting data may arise, leading to misinterpretation by agents and affecting subsequent processes [51]. Currently, there is a lack of algorithms capable of handling and resolving such conflicts.

**Computation Limit in Strategies**: A typical example is MAPF (Muti-agent path finding) which is a planning strategy. It aims to find paths for multiple agent systems, where the key constraint is that the agents will be able to follow these paths concurrently without conflicting with each other [52]. CBS (Conflict-Based Search) is most frequently used, leveraging hierarchical search. However, the huge search tree and constraint tree generated during CBS recursion are all stored in RAM, which has very high requirements on the server's hardware resources. In the future development of MAS to build a workflow, as the number of agents increases, the amount of data generated during CBS computation will grow exponentially so that the server's hardware resources will no longer support. The solution of the limitation can be combined with Reinforcement Learning, Direct Preference Optimization (DPO) and Tabu Search (TS) to record paths that are most likely to yield optimal solutions, as well as paths that could lead to stack overflow [4]. This method can improve the solution quality while helping to avoid deadlock issues in future computations.

*B. General Limitations*

A fundamental limitation lies in the lack of standardized specification mechanisms. This gap spans multiple dimensions. Most systems adopt self-defined DSLs or configuration formats, many of which consist of loosely structured prompt fragments. These lack formal syntax and semantics, making workflows hard to analyze, verify, or debug. There is no common modeling language to describe agent roles, workflows, and state transitions. Incompatible execution interfaces systems expose different execution entry points, planning conventions, and life cycle definitions, hindering cross-framework orchestration or reuse. Unlike in compiler or workflow engines, agent systems rarely share a unified intermediate representation (IR) or portable exchange format.

*C. Future Direction*

The core challenge and prospective development lies in the absence of standardized specification for building an agent workflow system. To address this, Google proposed open standards such as Agent2Agent (A2A) [76], enabling agents to communicate, share context, and collaborate seamlessly across platforms, regardless of provider or framework. This is a promising signal toward an industry-wide convergence to build a shared system for agent-based applications.

Also, we identify some key functionalities in Fig. 1. The emerging trends include multi-agent collaboration, dynamic planning, protocol standardization, and workflow optimization. Each direction is briefly illustrated and linked to relevant sections in this paper.

To scale further, workflows must support adaptive tool use, allowing agents to interface with APIs, search engines, databases, or computational modules in a context-aware manner.

Beyond extensibility, future systems must enable deeper customization ability to support complex, domain-specific tasks—ranging from academic research assistants to financial analysts—where agents coordinate reasoning, interaction, and memory in a task-specific configuration.

Multi-modal integration is rapidly becoming standard, enabling workflows to not only process language but also interact with images, code, documents, and structured data. At the same time, multi-agent collaboration frameworks are pushing the boundaries of coordination, particularly in decentralized, event-driven environments.

This lays the foundation for a new paradigm—systems that move from passive executors to autonomous, pervasive agents capable of long-term operation with minimal human input. To fully unlock the potential of agent workflows, the community must converge on unified framework and common abstractions that allow composability, interoperability, and modular deployment across platforms and providers. Only with such a foundation can we build a robust and scalable ecosystem that benefits research and industry alike.

## IX. CONCLUSION

In this survey, we systematically reviewed the emerging field of agent workflows. We examined both academic frameworks and industrial platforms.

We proposed two comparison analysis that evaluated systems across functionality dimensions and architectural and mechanism perspectives.

Throughout the survey, we emphasized the growing need for standardization, modularity, and orchestration capabilities. We highlighted trends in workflow modeling, workflow execution strategies and optimization.

We hope this article—and the collective exploration it inspires—can spark continued innovation and deeper systematization, helping agent workflows evolve from scattered practices into a unified foundation for building intelligent, goal-driven applications in the era of General Artificial Intelligence.

## ACKNOWLEDGEMENT

This work was funded by the National Natural Science Foundation of China (NSFC) under Grant [No.62177007].